\documentclass[conference]{IEEEtran}

\usepackage{mathptmx}
\usepackage[normalem]{ulem}
\usepackage{graphicx}
\usepackage{subfigure}
\usepackage[para,online,flushleft]{threeparttable}
\usepackage{caption}
\usepackage{booktabs}
\usepackage{amsmath}

\usepackage[bookmarks=true,breaklinks=true,letterpaper=true,colorlinks,linkcolor=black,citecolor=blue,urlcolor=black]{hyperref}

\usepackage{multirow}
\newcommand{\tabincell}[2]{\begin{tabular}{@{}#1@{}}#2\end{tabular}}

\begin{document}

\title{O-HAS: \uline{O}ptical \uline{H}ardware \uline{A}ccelerator \uline{S}earch for Boosting Both Acceleration Performance and Development Speed}

\author{\IEEEauthorblockN{Mengquan~Li, Zhongzhi~Yu, Yongan~Zhang, Yonggan~Fu, Yingyan~Lin$^\ast$}
\IEEEauthorblockA{Department of Electrical and Computer Engineering, Rice University, USA}
\IEEEauthorblockA{\{mengquan.li,~zy42,~yz87,~yf22,~yingyan.lin\}@rice.edu. \thanks{Corresponding author: Yingyan Lin. This article has been accepted to appear at ICCAD 2021.}}
}

\maketitle

\begin{abstract}

The recent breakthroughs and prohibitive complexities of Deep Neural Networks (DNNs) have excited extensive interest in domain specific DNN accelerators, among which optical DNN  accelerators are particularly promising thanks to their unprecedented potential of achieving superior performance-per-watt. However, the development of optical DNN accelerators is much slower than that of electrical DNN accelerators. One key challenge is that while many techniques have been developed to facilitate the development of electrical DNN accelerators, techniques that support or expedite optical DNN accelerator design remain much less explored, limiting both the achievable performance and the innovation development of optical DNN accelerators. To this end, we develop the first-of-its-kind framework dubbed O-HAS, which for the first time demonstrates automated 
\uline{O}ptical \uline{H}ardware \uline{A}ccelerator \uline{S}earch for boosting both the acceleration efficiency and development speed of optical DNN accelerators. Specifically, our O-HAS consists of two integrated enablers: (1) an O-Cost Predictor, which can accurately yet efficiently predict an optical accelerator's energy and latency based on the DNN model parameters and the optical accelerator design; and (2) an O-Search Engine, which can automatically explore the large design space of optical DNN accelerators and identify the optimal accelerators (i.e., the micro-architectures and algorithm-to-accelerator mapping methods) in order to maximize the target acceleration efficiency. Extensive experiments and ablation studies consistently validate the effectiveness of both our O-Cost Predictor and O-Search Engine as well as the excellent efficiency of O-HAS generated optical accelerators.

\end{abstract}
\section{Introduction}
\vspace{-0.5em}
With state-of-the-art inference accuracy, Deep Neural Networks (DNNs) have been widely employed in a myriad of AI applications, including speech recognition (e.g., Apple Siri and Amazon Alexa), face identification (e.g., Google Picasa), autonomous vehicles, etc. In parallel, the recent breakthroughs and prohibitive complexity of DNNs have excited extensive interest in domain specific DNN accelerators \cite{chen2016eyeriss,zhao2020smartexchange,9138916}. Among them, optical accelerators (OAs) provide a brand-new yet remarkably promising kind of computing platforms for DNNs. They can potentially achieve unparalleled massive parallelism, ultra-low latency, and little to no power consumption by leveraging silicon photonic technologies. It is estimated that OAs can be potentially 1000$\times$ faster than electronics with 1000$\times$ less power for the same die area \cite{nahmias2019photonic}. What's more, commercially manufacturable photonic integrated circuits (PICs) now also achieve economies of scale that is previously enjoyed solely by microelectronics \cite{shastri2021photonics}.

OA designs flourish since 2017, for which the capabilities of optics are widely explored to perform fast and efficient linear operations \cite{shen2017deep,liu2019holylight,xu202111,shiflett2020pixel,sunny2021crosslight}. Based on their core computing optical components and principles, they fall into three categories: First, Mach-Zehnder Interferometer (MZI)-based OAs were first proposed and could easily mass-fabricated for matrix-vector products using coherent lights; Singular value decomposition (SVD) \cite{shafiee2016isaac} and Fast Fourier transform (FFT) \cite{gu2020towards} implementations based on analog computing have been demonstrated, achieving high parallelism but suffering from a limited scalability to large-scale networks; Additionally, large-footprint MZIs also fail in on-chip integration. Second, to better tradeoff between the network scale and required footprint, an Electro-Optical Mach-Zehnder Modulator (EOM)-based OA \cite{xu202111} was presented by making full use of time, wavelength and spatial multiplexing with incoherent lights. Third, another alternative to MZI-based architectures, micro-resonator (MR)-based OAs \cite{liu2019holylight,shiflett2020pixel,sunny2021crosslight} are capable of operating across different wavelengths, modes, or polarization and occupying a much smaller chip area within a tight package, greatly increasing the compute parallelism, not only because many independent channels coexist in a single waveguide but also more optical components can be integrated on-chip.
\begin{figure*}[hbt]
\setlength{\abovecaptionskip}{0.1in}
\setlength{\belowcaptionskip}{-0.15in}
\centering
  \includegraphics[width=6.6in]{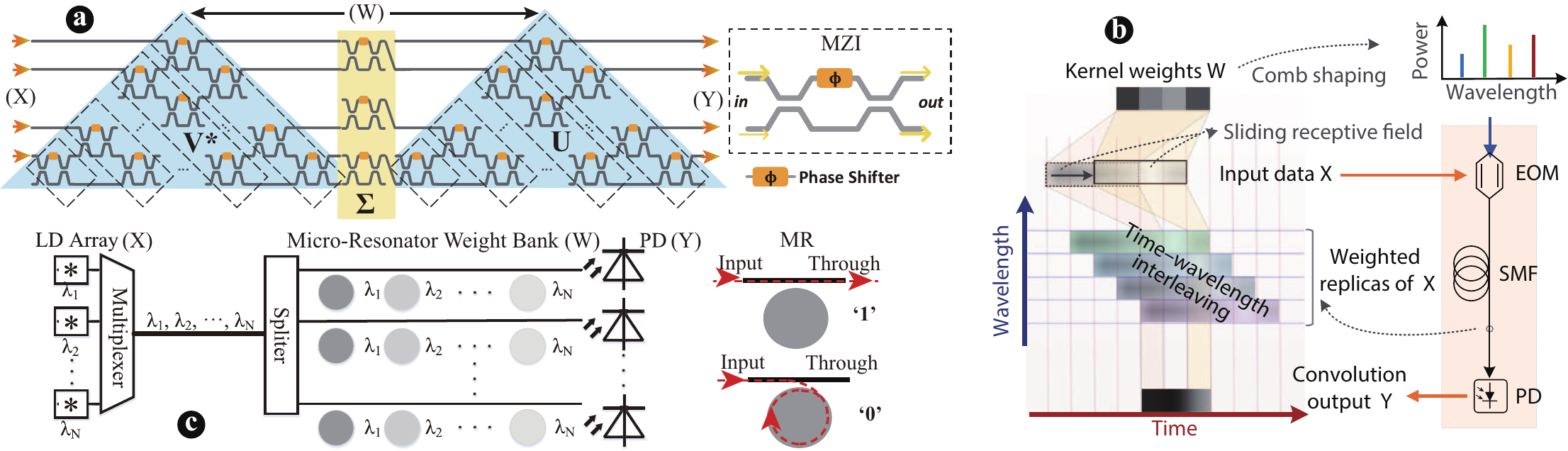}
    \caption{Three types of OCUs: a) a MZI-based optical SVD unit for matrix multiplication \cite{shen2017deep}; b) an EOM-based optical unit for vector convolution \cite{xu202111}; and c) a MR-based optical matrix-vector multiplier \cite{liu2019holylight}.}
  \label{f:OCUs}
\end{figure*}

\begin{figure}[!t]
\setlength{\abovecaptionskip}{0.1in}
\setlength{\belowcaptionskip}{-0.15in}
\centering
  \includegraphics[width=3.2in]{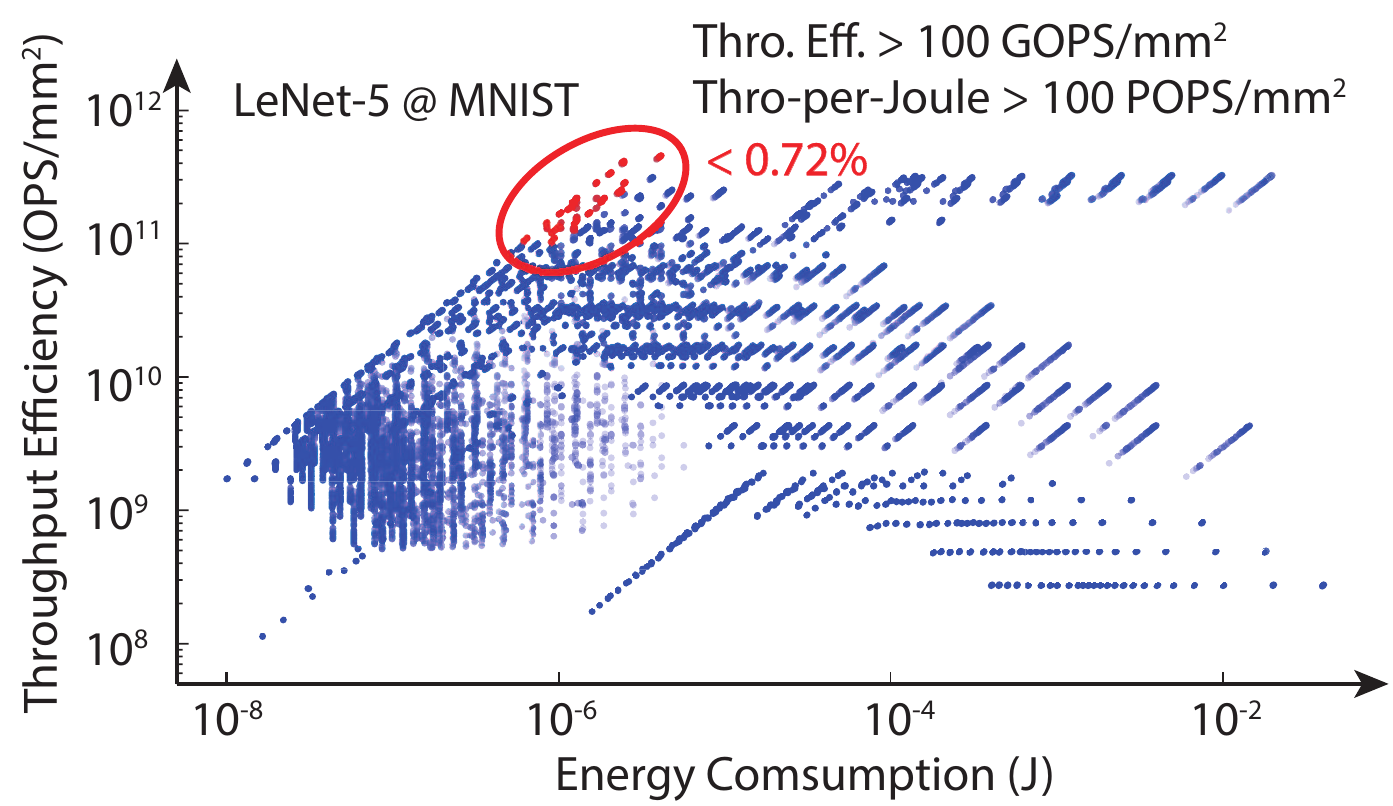}
  \vspace{-0.2em}
    \caption{OAs' performance on LeNet-5: There are a total of 1.42E+05 randomly sampled OA architecture points in this figure and the designs with \textit{Thro. Eff.} \textgreater 100 GOPS/$mm^2$ and \textit{Thro-per-Joule} \textgreater 100 POPS/$mm^2$ are marked as red, which are extremely sparse (\textless 0.72\%) within this search space.}
  \label{f:SearchSpace}
\end{figure}

However, the development of optical DNN accelerators is much slower than that of electrical DNN accelerators. Unlike the latter that many techniques have been developed to facilitate their development, techniques that support or expedite optical DNN accelerator design remain much less explored, limiting both the achievable performance and the innovation development of OAs. Overall, there are three critical bottlenecks: 1) Numerous OA architectures have been designed but no existing works fairly compare their advantages and shortages. For implementing the same DNN functionality, different OA architecture designs can lead to dramatically different performance/energy trade-offs. This makes it hard to obtain effective guidance for OA designers. 2) OA design is a cross-disciplinary field with a large design space, which often takes months to even years for manually designing effective OAs and requires cross-disciplinary knowledge, both limiting OAs' fast development. 3) In addition to the large design space formed by the extensive micro-architecture and dataflow choices, OA design is also application-specific. There has been no one-for-all OA design yet, which could perfectly suit massive and rapidly developed DNN models that vary in network shapes and sizes. Thus, ideally a customized OA architecture is required for a given DNN model and application pairs for achieving the best performance. This can be a prohibitive challenge for OA designers, especially when the ecosystem of facilitating OA development is currently still at its infancy. Moreover, as shown in Fig. \ref{f:SearchSpace}, satisfactory designs are extremely sparse in the large OA design space, which further exacerbates the difficulty for OA design.

 \vspace{0.5em}
To tackle the aforementioned challenges, we develop the first-of-its-kind framework dubbed \textbf{O-HAS} that for the first time demonstrates automated
\uline{O}ptical \uline{H}ardware \uline{A}ccelerator \uline{S}earch for boosting both the acceleration performance and development speed of optical DNN  accelerators. Our O-HAS consists of two integrated enablers: 1) an \textbf{O-Cost Predictor}, which can accurately yet efficiently predict an optical accelerator's energy and latency based on the DNN model parameters and the optical accelerator design; and 2) an \textbf{O-Search Engine}, which can automatically explore the large design space of optical DNN accelerators and identify the optimal accelerators (i.e., the type, shape, size of micro-architectures for both data computation and data access, algorithm-to-accelerator mapping methods, and memory hierarchy) in order to maximize the target acceleration efficiency. Extensive experiments and ablation studies consistently validate that the proposed O-Cost Predictor achieves excellent efficiency in not only the prediction accuracy (with an average error of $\sim$8.5\%) but also exploration and search time, which enables our O-Search Engine to perform fast and effective design space exploration. The O-HAS generated optical accelerators outperform the state-of-the-art (SOTA) optical accelerators by up to 119.70$\times$. 
\section{Prior Art and Opportunities}
\label{s:BM}

\textbf{OAs and Optical Convolution Units (OCUs).}
Current DNN models require dense linear operations such as convolutions or matrix computations, which accounts for the major computing costs in DNNs (e.g., 55\% to 90\% of the total computing power and time \cite{xu202111}). Optical linear operations exhibit stark advantages in bandwidth density, latency, and energy over their electrical counterparts. To boost inference efficiency, many optical micro-architectures have been designed for convolution acceleration \cite{shen2017deep,liu2019holylight,xu202111,shiflett2020pixel,sunny2021crosslight}. They are referred to as Optical Convolution Units (OCUs) in this paper. According to our comprehensive survey, we classify the SOTA OCUs into three categories as shown in Tab. \ref{t:OCU_para}, based on their operation principles and foundational computing optical components, i.e., Z-, E- and R-OCUs.

\uline{MZI-based OCUs (Z-OCUs):} They are the first type of OCUs proposed for matrix-vector multiplication acceleration using coherent light. SVD-based Z-OCU was designed and fabricated in \cite{shen2017deep}. It shows that every $N\times N$ weight matrix $W$ can be decomposed into three matrices using SVD, i.e., $W \overset{SVD}{=} U\Sigma V^*$, where $U$ and $V^*$ are $N\times N$ unitary matrices and $\Sigma$ is a diagonal matrix with singular values. MZIs are excellent candidates to optically implement unitary transformations as any arbitrary unitary matrix (e.g., $U$ and $V^*$) can be implemented with a triangular planar array of MZIs (shown in Fig. \ref{f:OCUs}(a)). By modulating the phase of the input signal via phase shifters, this OCU realizes matrix multiplication operations in the analog domain. SVD-based micro-architectures can achieve high computing speed close to light speed but is limited by their required large amount of hardware resources and thus heavy chip area cost. To improve the area efficiency over the SVD-based designs, a new FFT-based Z-OCU was further proposed in \cite{gu2020towards}. By leveraging optical fast Fourier transform and its inverse, it improved the area efficiency over SVD by $\sim$4$\times$ on average.

\uline{EOM-based OCUs (E-OCUs):} EOM-based OCU is another recently proposed micro-architecture for improving OAs' scalability and area efficiency \cite{xu202111}.
As shown in Fig. \ref{f:OCUs} (b), the input vector $X$ is encoded as the intensity of temporal symbols in a serial electrical waveform. The convolutional kernel is represented by a flattened weight matrix $W$ that is encoded to the optical power of the MicroComb lines via spectral shaping. The temporal waveform $X$ is then multi-casted onto the kernel wavelength channels via electro-optical modulation, generating the replicas weighted by $W$. Next, the optical waveform is transmitted through a dispersive single-mode fibre (SMF) with a delay step. Finally, the delayed and weighted replicas are summed via photodetector (PD). Each time slot yields a convolution between $X$ and $W$ for a given receptive field of convolution. The convolution window slides at the modulation speed to get the output vector $Y$. E-OCUs have the least number of optical devices and realize convolutions via effective time and wavelength interleaving. However, their heavy energy consumption is the major bottleneck.
\begin{table}[!t]
\setlength{\abovecaptionskip}{0in}
\setlength{\belowcaptionskip}{0in}
\footnotesize
\setlength{\tabcolsep}{1.1pt}
    \renewcommand{\arraystretch}{1.2}
        \caption{Hardware cost and performance comparison of SOTA OCUs when processing [1, $N$]$\times$[$N$, $N$] matrix-vector multiplication, where $B$ is the operand precision and $n_b$ is the multi-bit capability of MRs. For a fair comparison, the device counts are converted to \# MR based on real device footprints (the area and the energy ratio between one MZI/EOM and one MR are denoted as $\alpha$ and $\beta$, respectively ) \cite{gusqueezelight}.}
    \label{t:OCU_para}
    \centering
    \begin{threeparttable}
    \begin{tabular}{c|c|cc|ccc}
        \hline
        \multirow{2}{*}{} & \multirow{3}{*}{\tabincell{c}{\textbf{E-OCU}\\ \cite{xu202111}}} & \multicolumn{2}{c|}{\textbf{Z-OCU}}  &   \multicolumn{3}{c}{\textbf{R-OCU}}\\ \cline{3-7}
        &  & \tabincell{c}{SVD\\\cite{shen2017deep}}   &  \tabincell{c}{FFT\\\cite{gu2020towards}}  & \tabincell{c}{HolyLight\\\cite{liu2019holylight}} & \tabincell{c}{PIXEL\\\cite{shiflett2020pixel}} & \tabincell{c}{CrossLight\\\cite{sunny2021crosslight}} \\ \hline
        \# Wavelen.   & $N^2$B  & 1     & 1        & $\frac{B}{n_b}$ & B & $\frac{B}{\kappa n_k}$ \\
        Power     & $\beta$     & $\beta N(N-1)$  & $\sim\beta N(N-1)$ & $N^2\frac{B}{n_b}$ & $N^2$B & $N^2\frac{B}{n_b}$ \\
        Latency     & NB    & 1      & 1      & 1 & 1 & $\kappa$\\
        \tabincell{c}{Area\\(\# MR)} &$\alpha$  & $\alpha N(N-1)$ & $\sim\frac{\alpha}{4} N(N-1)$  & $N^2\frac{B}{n_b}$ & $2N^2B$ & $N^2\frac{B}{n_b}$ \\
        \tabincell{c}{Control\\Complexity}  & Low  & \multicolumn{2}{c|}{High} & \multicolumn{2}{c}{Medium} & High \\ \hline
\end{tabular}
\begin{tablenotes}
*For R-OCUs, there is a trade-off between latency and area cost as matrix multiplication can be performed by either spatial multiplexing with multiple R-OCUs or time multiplexing on the same R-OCU. In this table, we show the cases with the optimal latency via spatially-multiplexed architectures.
\vspace*{-0.1in}
\end{tablenotes}
\end{threeparttable}
\end{table}

\uline{MR-based OCUs (R-OCUs):} Compared with MZIs and EOMs, MRs (such as microrings and microdisks) have much smaller footprints and thus are excellent candidates for implementing photonic integrated circuits (PICs). R-OCUs have attracted the most attention for on-chip OA designs \cite{liu2019holylight,shiflett2020pixel,sunny2021crosslight}. A typical MR-based OCU relies on an array of on-chip laser sources, MRs, PDs, a wavelength splitter, and a multiplexer to conduct parallel matrix multiplication operations, as shown in Fig.~\ref{f:OCUs}(c). To calculate the multiplication between matrix $W$ and vector $X$, elements in matrix $W$ are represented by the transmissivity of an $N \times N$ MR array, while elements in vector $X$ are denoted by the input optical power values produced by $N$ on-chip laser diodes (LDs) with different wavelengths $(\lambda_1, \cdots,\lambda_N)$. Vector $X$ is multiplexed, split, and transmitted to each row of the matrix equally. An MR with a specific resonance wavelength in a row of the matrix will react on the input signal in the same wavelength, and the output power is then the product between the input power and the MRs' transmissivity. In a row, $N$ multiplications between the row vector of $W$ and the vector $X$ are realized and then aggregated together in the same waveguide. The multiply-accumulate (MAC) result is finally collected by the PD. The same operations occur in all rows, thus the R-OCU finishes all convolutions with high parallelism. We have listed three SOTA R-OCUs in Tab. \ref{t:OCU_para}. They achieve different performances resulting from different manufacturing technologies and architecture parameters.

\textbf{Opportunities.} Although these OAs are promising with exciting performance, OA development is at its very early stage. All of them are manually designed with a great deal of design/time overhead. There lacks a mature ecosystem to support and assist OA designers and researchers, which largely limits the development of this new field. To resolve this problem, there are two necessities. One is a generic predictor that enables accurate yet efficient estimation of the performance and hardware cost of any arbitrary OAs. It is vital because this predictor makes it practical to fairly compare the pros and cons of distinct OAs, which allows OA designers to accurately analyze accelerators' performance and hardware cost before the actual implementation, thus can largely reduce the design cycles and facilitates the development of OAs. The other essential is an automated search framework for effective optical DNN accelerators exploration. Considering the vast and increasing gap between the prohibitive complexity of powerful DNN models and OA development speed, this framework should perform automated exploration of OAs' large design space and generate the optimized designs satisfying the target acceleration efficiency, without humans in the loop. To sum up, both factors are indispensable to boost both OAs' achieved acceleration performance and development speed. In this paper, we present the first-of-its-kind framework dubbed O-HAS that for the first time demonstrates automated optical accelerator search.
\begin{figure}[!t]
\setlength{\abovecaptionskip}{0.1in}
\setlength{\belowcaptionskip}{-0.15in}
\centering
  \includegraphics[width=3.5in]{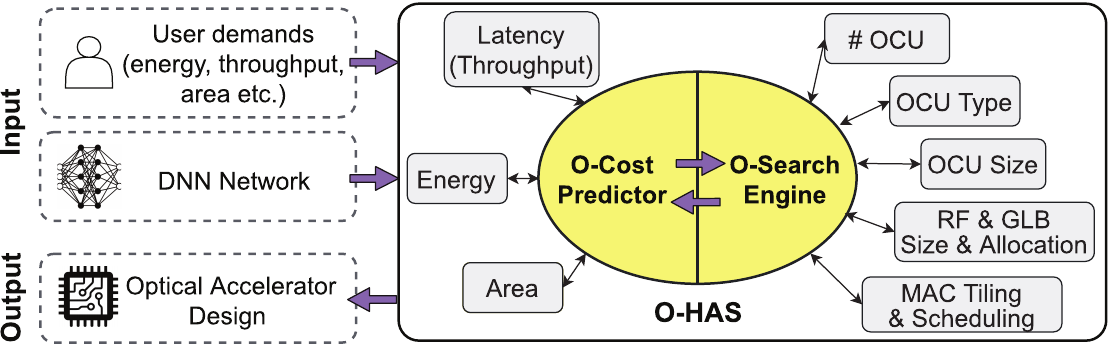}
  \caption{The proposed O-HAS framework.}
  \label{f:OHAS}
\end{figure}
\begin{figure*}[hbt]
\setlength{\abovecaptionskip}{0.05in}
\setlength{\belowcaptionskip}{-0.15in}
\centering
  \includegraphics[width=6.8in]{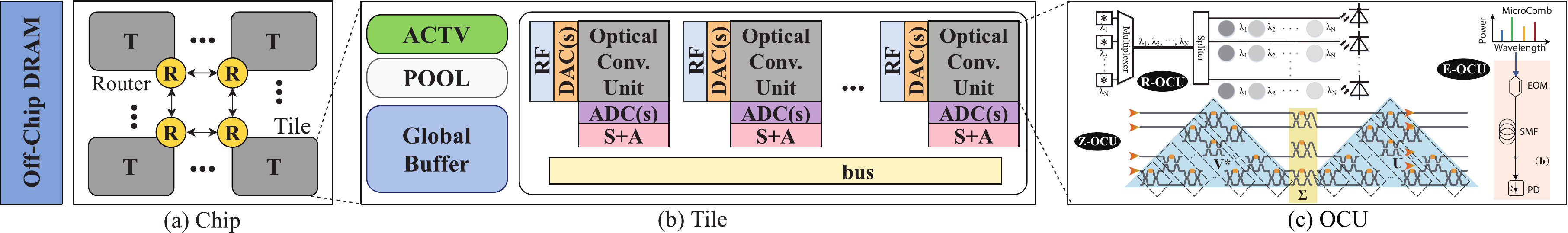}
  \caption{An illustration of generic optical accelerator architectures.}
  \label{f:OAA}
\end{figure*}

\section{The Proposed O-HAS framework}
\label{s:O_HAS}

To automatically search for the optimal hardware architecture for a given DNN model, we first unify the taxonomy of a generic optical accelerator architecture and specify the huge search space, in order to ensure a wide applicability.
After that, we develop the \textbf{O-HAS} framework as illustrated in Fig. \ref{f:OHAS}. O-HAS integrates two enablers: 1) an O-Cost Predictor, to accurately yet efficiently predict an optical accelerator's energy, latency (as well as throughput), and area costs based on the DNN model parameters and the optical accelerator design; and 2) an O-Search Engine, to automatically explore the large design space of OAs and identify the optimal accelerators that maximize the target user-defined acceleration efficiency.

\subsection{Generic Optical DNN Accelerator}
\textbf{Overview.} As shown in Fig. \ref{f:OAA}, a generic optical accelerator includes an off-chip DRAM and a chip node, the latter of which consists of multiple tiles connected via an on-chip network. Each tile communicates with the others through its router, and relies on a global buffer (GLB) and multiple register files (RFs) to store and load the operands and intermediate results generated by the computation units. As the main computation components, OCU, POOL, and ACTF units are used for the convolution, pooling, and activation operations, respectively. Convolutions during DNN inferences can be computed by multiple OCUs connected through a shared bus. For each OCU, one or multiple DAC(s) and ADC(s) (depending on their sampling frequency) and electrical shift-and-add (S+A) units are used for digital-analog conversion and aggregation, respectively. In addition to the OCUs, POOL, ACTV, S+A circuits, DACs and ADCs can also be potentially implemented with optics, which are far from mature yet \cite{wetzstein2020inference}. As such, We focus on OCUs in this paper and leave the other components to future works.
\begin{figure}[t!]
\setlength{\abovecaptionskip}{0.05in}
\setlength{\belowcaptionskip}{-0.2in}
\centering
  \includegraphics[width=3.3in]{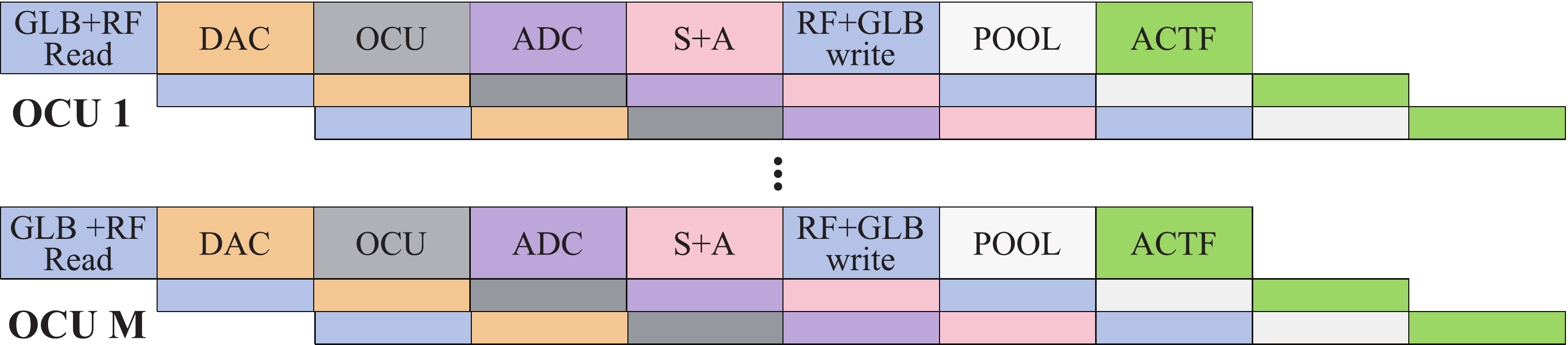}
  \caption{The optical convolution unit pipeline.}
  \label{f:pipeline}
\end{figure}

\textbf{Computing flow and pipeline.} Different types of OAs have different dataflows. For accelerators using R- and Z-OCUs, after a DNN model is sufficiently trained, weight matrices of various layers of the DNN will be pre-programmed into OCUs by configuring the transmissivity of the resonators in each R-OCU and the transmission phase angle of the phase shifters in each Z-OCU. on the other hand, different from R- and Z-OAs, the weight matrices of the DNN will not be stationed in E-OCUs, with no local reuse. As illustrated in Fig. \ref{f:pipeline}, for an OA, inputs of a CONV layer are pre-fetched from off-chip DRAMs and routed to available tiles and stored in their global buffers. During inferences, the input pixels are tiled and feed to OCUs' register files from the global buffer. They will be converted into optical signals via DAC(s) and electrical-optical (E/O) conversion in OCUs (e.g., using optical transmitters (TX) like LDs and MicroCombs). To collaboratively compute convolutions, the accelerator controller steers OCUs producing intermediate results in the form of electrical voltages via optical-electrical (O/E) conversion in OCUs (e.g., using optical receiver (RX) like PDs). The intermediate result voltages are sampled and converted to digital values by ADCs. Digital results are aggregated in register files and global buffers via S+A operations, and sent to pooling and activation units to generate layer outputs that will be buffered back to the global buffer for processing the next layer. The computation continues until the final result is produced. It is noteworthy that the processing among multiple layers in one OCU will be executed in a pipeline manner, while multiple OCUs are able to perform in parallel to boost the throughput. Here we simplify the pipeline and assume the number of cycles occupied by each stage to be the same in Fig. \ref{f:pipeline}, which actually depends on the execution latency of each stage. A detailed latency analysis will be presented in Section \ref{s:predictor}.
\begin{table}[b!]
\small
\setlength{\abovecaptionskip}{0in}
\setlength{\belowcaptionskip}{-0.1in}
\setlength{\tabcolsep}{1.3pt}
    \renewcommand{\arraystretch}{1.2}
    \caption{Generic optical DNN accelerator search space.}
    \label{t:search_space}
    \centering
    \begin{tabular}{cccc||cc}
        \hline
        \multicolumn{2}{c}{\textbf{Memory Hierarchy}} &  \multicolumn{2}{c}{\textbf{Loop-order}} &  \multicolumn{2}{c}{\textbf{Loop-tiling}} \\\hline
        \multicolumn{2}{c}{Register File (RF)} & \multicolumn{2}{c}{$\surd$}  & \multicolumn{2}{c}{$\surd$} \\
        \multicolumn{2}{c}{Global Buffer (GLB)} & \multicolumn{2}{c}{$\surd$}  & \multicolumn{2}{c}{$\surd$} \\
        \multicolumn{2}{c}{DRAM} & \multicolumn{2}{c}{$\surd$} & \multicolumn{2}{c}{--}\\ \hline \hline
        \textbf{Computation} & \textbf{Type} & \textbf{\# of units} & \textbf{Size of units} & \textbf{Perip.} & \textbf{\# of Tile}\\ \hline
        OCU & $\surd$ & $\surd$ & $\surd$ & Tile &$\surd$ \\ \hline
\end{tabular}
\end{table}

\textbf{Search space definition.}
Similar to Network Architecture Search (NAS), an search space of optical hardware accelerators is a prerequisite for accelerator architecture exploration and optimization. However, it is challenging to specify such a space for DNN optical accelerators due to their huge design space. First, while the capabilities of optics to efficiently perform linear convolutions have been widely explored, there are numerous optical devices that can realize convolutions, such as MRs, EOMs, and MZIs. Different devices have their unique optical responses and characteristics, resulting in distinct performance and hardware costs when being used to build accelerators.
Second, there are lots of ways to design the accelerators' micro-architectures, which are characterized by the number of hierarchical memory (e.g., RF, GLB, and DRAM) and computation units (e.g., tile, OCU, POOL, and ACF), the size of each memory level, the type and size of computation units (e.g., there are 3 types of OCUs for OAs), and the NoC design \cite{chen2016eyeriss}. Third, there are a lot of choices for the algorithm-to-hardware mapping methods, i.e., how to temporally and spatially schedule all the operations in DNNs to be executed in the target accelerators, with fine-grained consideration for the loop order of the memory hierarchy as well as MAC tiling and scheduling.

We extract a search space for generic optical DNN accelerators in Tab. \ref{t:search_space} by leveraging the nested for-loop accelerator description. Here the descriptions of the loop-order and loop tiling within each memory level are referred to \cite{parashar2019timeloop}, which aims to maximize local data reuses for optimal memory access performance and minimal data access cost. A similar methodology is utilized in our O-Search Engine. In addition, different types of OCUs will occupy distinct chip areas and consume different time/energy costs, and there are numerous choices for the sizes and numbers of OCUs.

\subsection{The O-Cost Predictor}
\label{s:predictor}
\begin{figure}[tb!]
\setlength{\abovecaptionskip}{0.05in}
\setlength{\belowcaptionskip}{-0.2in}
\centering
  \includegraphics[width=3.1in]{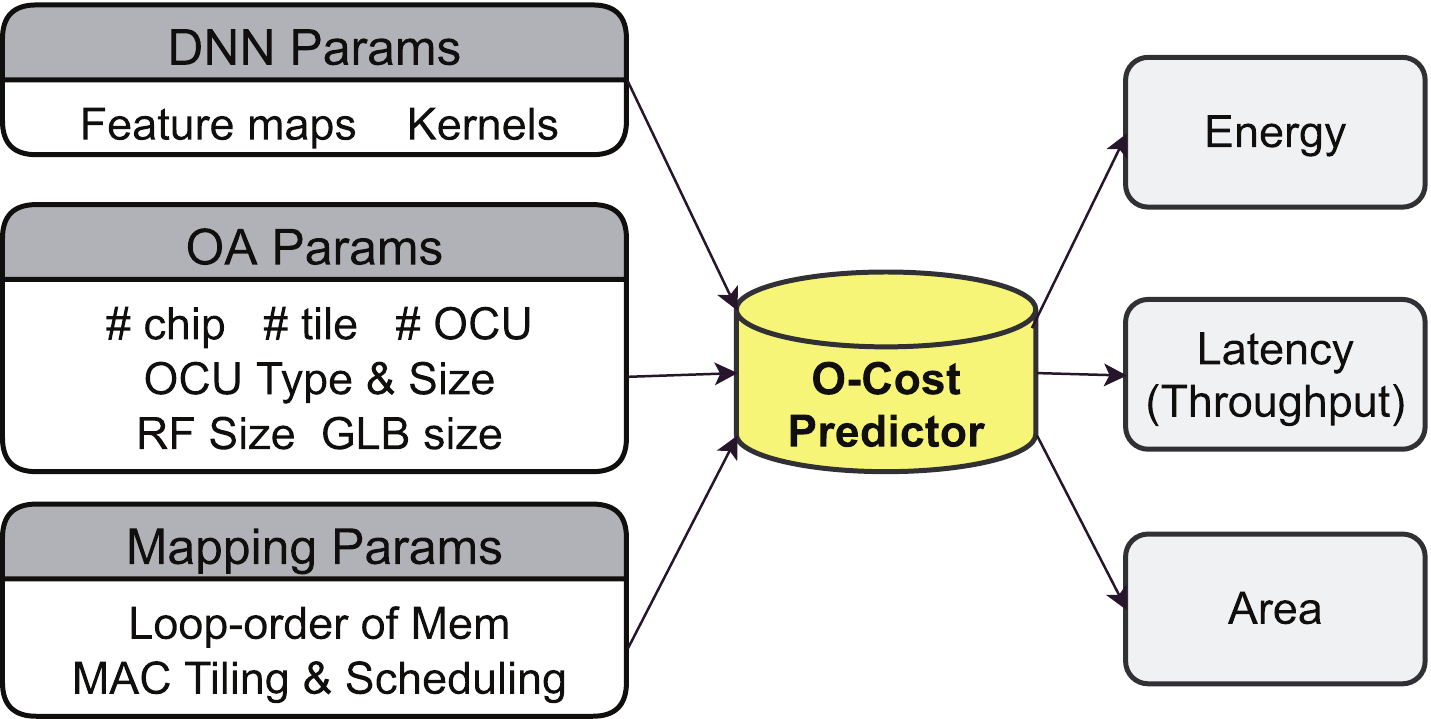}
  \caption{The proposed O-Cost Predictor.}
  \label{f:O_Predictor}
\end{figure}

Based on the generic OA architecture space presented above, we then develop an O-Cost Predictor as shown in Fig. \ref{f:O_Predictor} to evaluate the latency (as well as throughput), the energy consumption, and the area cost of an given optical DNN accelerator, by jointly considering algorithmic (e.g., DNN parameters including the height and weight of input and output feature maps, kernels, input and output channels), architectural (e.g., the number of tiles, the type, number and size of OCUs, the size of RFs and GLBs), and mapping level (e.g., loop-order of memory hierarchy, MAC tiling and scheduling) parameters.

\textbf{Energy.} Modern DNNs usually consist of a cascade of multiple convolution (CONV), fully-connected (FC), pooling, normalization, and activation function layers, through which the inputs are progressively processed. In this paper, we focus on the energy consumed by convolutions as it is the major contributor to the total energy consumption. Given a DNN model that contains $J$ CONV/FC layers, the overall energy consumption can be modeled as:
\vspace{-0.08in}
\begin{equation}\small
  E = \sum^J_{i=1}E_{mem}^{i} + E_{comp}^{i}
  \vspace{-0.08in}
  \label{e:E_DNN}
\end{equation}
where $E_{mem}^{i}$ and $E_{comp}^{i}$ are the energy costs of data access and data computation in the $i$-th CONV/FC layer, respectively. An FC layer's energy cost can be similarly modeled.

Different types of OCUs will result in distinct hardware costs in energy, latency, and area. In this section, we take R-OCUs as an example due to the space limit, the principle of which can be applied to modeling E- and Z-OCUs. Parameters used in our Predictor are listed in Tab. \ref{t:model_para}.

\uline{Data Access Energy:} As kernel weights stay stationary in one or multiple R-OCU(s), the data access energy of one layer is composed of the data movement energy of the input pixels, intermediate results (\emph{Psums}), and output pixels. Their access energy is proportional to the unit energy per data access and the total number of data accesses. Each input is pre-read from the GLB in the tile to the RF in the OCU, and then modulated into different wavelengths via DAC(s) and TX(s) for computation. If all the weights in one layer is mapped onto a $N\times N$ OCU, \emph{Psums} are only written to and read from its RF, of which the energy per Psum access can be denoted as ($e^{RF}_{read} + e^{RF}_{wrt}$). Otherwise, if one layer is large and needs to be mapped onto multiple OCUs, \emph{Psums} will also be accessed from the GLB in addition to the RFs, for which the energy per Psum access can be denoted as ($e^{GLB}_{read} + e^{GLB}_{wrt}$). The output write energy of this layer is proportional to the number of outputs  written into the GLB (i.e., $E_iF_iD_i$) and the energy per output write (i.e., $e^{GLB}_{wrt}$).
\begin{equation}\small
\label{e:R_E_mem}
E_{mem}^{i}=\left\{
\begin{array}{lc}
(e^{GLB}_{read} + e^{RF}_{read})E_iF_iZ^2_iC_i\lceil\frac{D_iN_b}{N}\rceil+& (E_{input})\\
(e^{RF}_{read} + e^{RF}_{wrt})E_iF_iD_iN_b\lceil\frac{Z^2_iC_i}{N}\rceil+& (E^{RF}_{Psum})\\
(e^{GLB}_{read} + e^{GLB}_{wrt})E_iF_iD_iN_b\times\\
\lceil\lceil\frac{Z^2_iC_i}{N}\rceil\lceil\frac{D_iN_b}{N}\rceil/(K_tK^{Ty}_{OCU}) -1\rceil+& (E^{GLB}_{Psum})\\
e^{GLB}_{wrt}E_iF_iD_i& (E_{output})
\end{array} \right.
\end{equation}
\begin{table}[tb!]
\footnotesize
\setlength{\abovecaptionskip}{0in}
\setlength{\belowcaptionskip}{0in}
\setlength{\tabcolsep}{1.0pt}
    \renewcommand{\arraystretch}{1.1}
    \caption{Parameters used in the O-Cost Predictor.}
    \label{t:model_para}
    \centering
    \begin{tabular}{|c|c|}
        \hline
        \textbf{DNN Params} & \textbf{Description}\\ \hline
        $C_i$/$D_i$ & input/output channel in the i-th layer \\ \hline
        $H_i$/$W_i$ & input feature map height/weight in the i-th layer \\ \hline
        $Z_i$/$S_i$ & filter height (= weight)/stride in the i-th layer \\ \hline
        $E_i$/$F_i$ & output feature map height/weight in the i-th layer \\ \hline\hline

        \textbf{OA Params} & \textbf{Description}\\ \hline
        $K_{t}$, $K^{Ty}_{OCU}$ & \# of tiles in one accelerator; \# of OCUs in one tile  \\ \hline
        $Ty$ & \tabincell{c}{Type of OCUs in one tile (Type R, E, Z in this paper)} \\ \hline
        $e^{RF}_{read}$/$e^{RF}_{wrt}$ & the unit energy of the register file read/write \\ \hline
        $e^{GLB}_{read}$/$e^{GLB}_{wrt}$ & the unit energy of the global buffer read/write \\ \hline
        $e^D_{read}$/$e^D_{wrt}$ & the unit energy of the DRAM read/write \\ \hline
        $l^{RF}$, $l^{GLB}$ & the absolute hit latency of register file and global buffer access \\ \hline
        $l^{DRAM}$ & the absolute hit latency of the DRAM access \\ \hline
        $Q^{RF}$, $Q^{GLB}$ & the size of the register file and global buffer \\ \hline
        $Q^{DRAM}$ & \# of data transferred from the off-chip DRAM \\ \hline
        $e_{TX}$, $e_{RX}$ & \tabincell{c}{the unit energy of the optical transmitter and receiver} \\ \hline
        $e_{R}$ & \tabincell{c}{the unit energy per symbol of a MR (w/ multi-bit capacity)} \\ \hline
        $e_{tune}$ & the unit tuning energy of a MR \\ \hline
        $e_{D/A}$, $e_{A/D}$ & the unit energy of the D/A and A/D conversion \\ \hline
        $e_{S+A}$ & the unit energy of the Shift-and-Add circuits \\ \hline
        $R_{R}$ & the symbol rate of a MR \\ \hline
        $N_b$, $B$& \# of MRs for mapping one weight; \# of bits per weight \\ \hline
        $N\times N$, $a_{R}$ & the size of an OCU; the area of a R-OCU \\ \hline
        $a^{RF}$/$a^{GLB}$ & the unit area of the register file and global buffer\\ \hline
\end{tabular}
\vspace*{-0.1in}
\end{table}

\uline{Data Computation Energy:} The data computation energy of one layer consists of the energy consumed by the D/A and A/D conversions, E/O and O/E conversions using TXs and RXs, MAC, local tuning, and S+A operations, i.e.,:
\begin{equation}\small
\label{e:R_E_comp}
E_{comp}^{i}=\left\{
\begin{array}{lc}
(e_{D/A}+e_{TX})E_iF_iZ^2_iC_i\lceil\frac{D_iN_b}{N}\rceil+& (E_{D/A})\\
(e_{R}+e_{tune})E_iF_iD_iN_bZ^2_iC_i+ & (E_{MAC})\\
(e_{RX}+e_{A/D}+e_{S+A})E_iF_iD_iN_b\lceil\frac{Z^2_iC_i}{N}\rceil & (E_{A/D})
\end{array} \right.
\end{equation}

\textbf{Latency and Throughput.}
Considering that the accelerators can operate in a layer-wise pipeline manner with fine-grained balanced pipeline as shown in Fig. \ref{f:pipeline}, the OA throughput and the latency can be modeled as (\ref{e:T_DNN}):
\begin{equation}\small
\label{e:T_DNN}
\begin{array}{c}
  T =  \frac{\# MACs}{max(L^{1}, L^{2}, ..., L^{J})} \quad\&\quad L^{i} = max(L^i_{mem}, L^i_{comp})
 \end{array}
\end{equation}
where $L^i_{mem}$ and $L^i_{comp}$ denote the data access latency and data computation latency of the $i$-th layer, respectively.

\uline{Data Access Latency:} The total access time for the RF, GLB and DRAM can be modeled as (\ref{e:L_mem}). Given the size of RF and GLB, each access is either a hit or a miss. Considering the three-level memory hierarchy in this paper, we can calculate the average data access time as the sum of the time spent in hits and that spend on misses. We calculate the absolute hit rate as the fraction of all accesses that hit in this level of memory. Therefore, assumed that the requirements for RF and GLB access to generate one 3D output pixel is respectively $Q^{RF}_{req}$ and $Q^{GLB}_{req}$, we can calculate the absolute hit rate for RF and GLB accesses as (\ref{e:L_mem}), and thus obtain the total data access time. Note that there are two contributors to GLB access. One is from RF access misses, and the other is from \emph{Psums} accesses due to inter-OCU data movement.
\begin{equation}\footnotesize
\label{e:L_mem}
L_{mem}^{i}=E_iF_i\left\{
\begin{array}{lc}
l^{RF}\cdot min(\frac{Q^{RF}K_tK^{Ty}_{OCU}}{Q^{RF}_{req}},~1) + & L^{RF}_{Access}\\
l^{GLB}\cdot\big(\frac{min(Q^{GLB},~ Q^{RF}_{req}-Q^{RF}K_tK^{Ty}_{OCU})}{Q^{RF}_{req}} + min(\frac{Q^{GLB}K_t}{Q^{GLB}_{req}},~1)\big) + & L^{GLB}_{Access} \\
l^{DRAM}\cdot\Big(\frac{min(Q^{DRAM},~ Q^{RF}_{req}-Q^{RF}K_tK^{Ty}_{OCU}-Q^{GLB}K_t)}{Q^{RF}_{req}} +\\
\qquad\qquad\frac{min(Q^{DRAM},~Q^{GLB}_{req}-Q^{GLB}K_t)}{Q^{GLB}_{req}}\Big) & L^{DRAM}_{Access}\\
\end{array} \right.
\end{equation}

\uline{Data Computation Latency:} Every MR in a R-OCU achieves a transmission rate of $R_R$. Assuming that there are $K^{Ty}_{OCU}$ R-OCU(s) per tile and $K_t$ tile(s) on-chip in total, it takes $\lceil\frac{\lceil Z^2_iC_i/N\rceil\lceil D_iN_b/N\rceil}{K_tK^{Ty}_{OCU}}\rceil$ time cycles to generate one 3D output pixel. Thus, to generate $E_iF_i$ 3D output pixels, the computation latency can be formulated as follow:
\begin{equation}\small
  L^{i}_{comp} = E_iF_i\lceil\lceil\frac{Z_i^2C_i}{N}\rceil\lceil\frac{D_iN_b}{N}\rceil/(K_tK^{Ty}_{OCU})\rceil/R_{R}
\end{equation}

\textbf{Area Cost.}
We model the area cost of OAs by taking into account the area for both the data access and convolution computation:
\begin{equation}\small
\label{e:A_DNN}
\begin{array}{c}
  A = a^{RF}Q^{RF}K_tK^{Ty}_{OCU}+ a^{GLB}Q^{GLB}K_t + a_{R}K_tK^{Ty}_{OCU}
\end{array}
\end{equation}

\begin{figure}[tb!]
\setlength{\abovecaptionskip}{0.05in}
\setlength{\belowcaptionskip}{-0.15in}
\centering
\subfigure[Energy prediction accuracy.]
{\includegraphics[width=0.83\columnwidth]{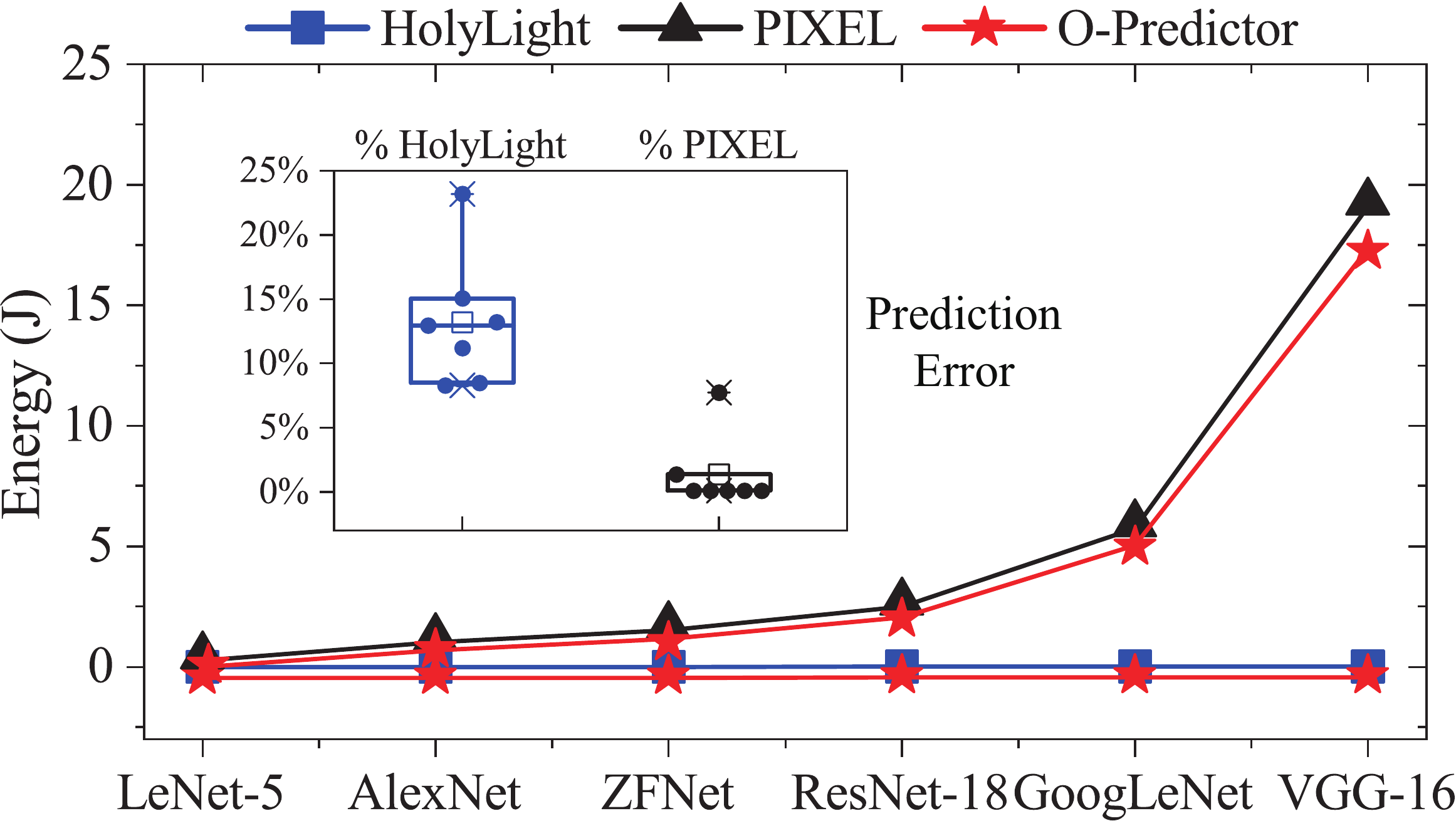}}
\subfigure[Latency prediction accuracy.]
{\includegraphics[width=0.83\columnwidth]{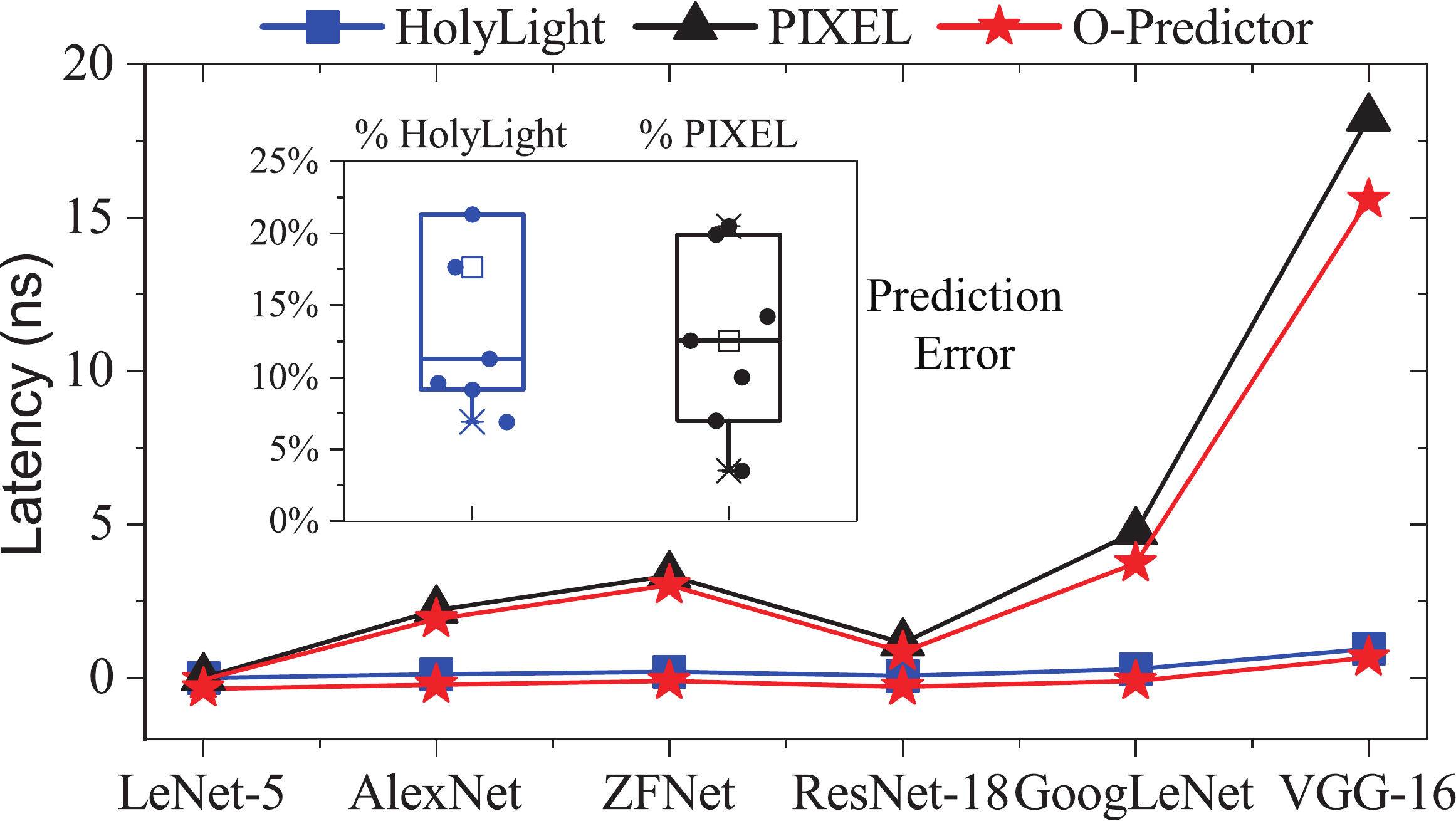}}
\caption{The prediction accuracy of our O-Cost Predictor against HolyLight and PIXEL, where the error boxplots show the statistical distribution of the prediction errors ($Error$ = $|V_{Pred} - V_{GT}|/V_{GT}$, while $V_{Pred}$ and $V_{GT}$ are our O-Cost Predictor's estimated value and the ground truth result, respectively).}
\label{f:OPred_Accuracy}
\end{figure}

\textbf{O-Cost Predictor Validation.}
We validate the accuracy of the proposed O-Cost Predictor in energy and latency prediction against two SOTA OAs (i.e., HolyLight \cite{liu2019holylight} and PIXEL \cite{shiflett2020pixel}). In particular, we evaluated the O-Cost Predictor on six DNNs (including LeNet-5, AlexNet, ZFNet, ResNet-18, GoogLeNet, and VGG-16), and compare the predicted results with the results obtained from HolyLight and PIXEL (which are set as the ground truth values). We use a deep learning accelerator simulator FODLAM \cite{fodlam} for HolyLight and an in-house simulator for PIXEL. As shown in Fig. \ref{f:OPred_Accuracy}(a), the energy prediction error between our Predictor and the ground truth values in HolyLight are between 2\% and 17\%, and the differences between our Predictor and PIXEL are between 5\% and 7\%. These errors are acceptable with an average of $\sim$8.5\% while supporting generic MR-based optical accelerators.

The latency prediction is much more complex than energy estimation, as it is hard to accurately emulate the dataflow and the pipeline control of accelerators. We adopt the same dataflow and the pipeline strategy in HolyLight and validate the latency model of the proposed O-Cost Predictor against HolyLight and PIXEL. Fig. \ref{f:OPred_Accuracy}(b) shows the results. The latency prediction error between our Predictor and the ground truth values in HolyLight are between 9\% and 23\%, and the differences between our Predictor and PIXEL are between 7\% and 14\%, with a satisfactory average error of $\sim$12.5\%.

\subsection{The O-Search Engine}
To realize a generic OA design search engine, our O-Search Engine aims to solve the OA design parameter $\gamma^*$ that satisfies
\begin{equation}\small
    \gamma^* = arg\min_\gamma L_{hw}(NET, HW(\gamma)),
    \label{eq:opt goal}
\end{equation}
where $NET$ is the given DNN architecture and $HW$ is the OA design search space parameterized by $\gamma$, $L_{hw}$ denotes the resulting hardware cost for the $NET$ and $HW(\gamma)$ pair.

To achieve the optimal trade-off between the exploration and exploitation during search, we reformulate Eq. (\ref{eq:opt goal}) to the following format:
\vspace{-0.05in}
\begin{equation}\small
    \gamma^* = arg\min_\gamma \sum_{i=1}^S L_{hw}(NET, HW(GS(\gamma^1),\cdots,GS(\gamma^S))),
    \vspace{-0.05in}
\end{equation}
where $\gamma^i(i=1,\cdots,S)$ is a normalized vector representing the $i$-th OA design parameter, the $j$-th value of $\gamma^i$ represents the probability of selecting the $j$-th choice to be the OA's $i$-th parameter, $S$ is the total number of design parameters in the OA design search space, $GS(\gamma^i)$ is the Gumbel-Softmax~\cite{gumbel1954statistical,maddison2014sampling} sampling of the $i$-th parameter parameterized by $\gamma^i$ from the OA design search space.
In each update, we apply Gumbel-Softmax to each searchable parameter in the OA design search space and sample one OA design, and then evaluate the hardware cost $L_{hw}(NET, HW(GS(\gamma^1),\cdots,GS(\gamma^S)))$ of the sampled OA design using our proposed O-Cost Predictor in Eq.~\ref{s:predictor}.
After that, $L_{hw}(NET, HW(GS(\gamma^1),\cdots,GS(\gamma^S)))$ is multiplied with the sampled $GS(\gamma^i)$, for $i=1,\cdots,S$ to relax $\gamma^i(i=1,\cdots,S)$ during backpropogation.
In this manner, the randomness of Gumbel-Softmax helps the exploration of the whole search space while the relaxation of $\gamma^i(i=1,\cdots,S)$ helps the exploitation of the optimal OA design. 
\section{Performance Evaluation}

\textbf{Benchmark DNNs and Datasets.} We studied our O-HAS framework on six DNNs including LeNet-5, AlexNet, ZFNet, ResNet-18, GoogLeNet, and VGG-16. LeNet-5 was trained with MNIST to classify simple handwritten digits, while the rest were trained with ImageNet to recognize complex objects. More details of the network topology are shown in Tab. \ref{t:bench_topo}.
\begin{table}[b!]
\setlength{\abovecaptionskip}{0in}
\setlength{\belowcaptionskip}{-0.15in}
\centering
\footnotesize
\caption{DNN benchmarks \cite{liu2019holylight, thop}.}
\setlength{\tabcolsep}{5pt}
\begin{tabular}{|c|c|c|c|}\hline
\textbf{Name} & \textbf{Data Set} & \textbf{Topology} & \textbf{Total \# MACs}\\\hline
LeNet-5                & MNIST                      & 3C,2P,2F       & 2.86E+05   \\\hline
AlexNet                & ImageNet                   & 5C,3P,3F       & 7.15E+08     \\\hline
ZFNet                  & ImageNet                   & 5C,3P,2F       & 7.78E+08 \\\hline
ResNet-18               & ImageNet                  & 20C,2P,1F      & 1.82E+09 \\\hline
GoogLeNet               & ImageNet                  & 57C,14P,1F       & 1.50E+09 \\\hline
VGG-16                  & ImageNet                  & 13C,6P,3F     & 1.55E+10     \\\hline
\end{tabular}
\label{t:bench_topo}
\end{table}

\textbf{Methodology.} We build the O-Search Engine in PyTorch for automated design space exploration, in which we use the Adam~\cite{kingma2014adam} optimizer with a learning rate of $1\times 10^{-7}$, $\beta_1=0.5$, and $\beta_2 = 0.999$. For each DNN benchmark, we obtained two optimized OA designs using our O-Search Engine, subjecting to the chip area constraint. These two designs, which are referred to as OHAS-EA and OHAS-TEA in this paper, prioritize high energy efficiency and throughput-per-Joule performance, respectively. We then studied their accelerator performance and inference performance for practical uses, by comparing them against six SOTA OAs, including EOM \cite{xu202111}, SVD \cite{shen2017deep}, FFT \cite{gu2020towards}, HolyLight \cite{liu2019holylight}, PIXEL \cite{shiflett2020pixel}, and CrossLight \cite{sunny2021crosslight}).

\textbf{Accelerator modeling.} We use our O-Cost Predictor to study the energy consumption, throughput, and area cost of all the accelerators running different DNN models. For a fair comparison, we are consistent in the parameters for the core functional components used in all the accelerators. All the parameters of the optical devices are generated based on a 130 nm SOI CMOS process, and the parameters of hierarchical memories are obtained based on a commercial 28 nm process technology. We model and adopt On-chip LDs from \cite{stern2018battery}, MR from \cite{zheng2012ultralow}, optical Splitter\&MUX from \cite{subbaraman2015recent,hosseini20111}, PDs from \cite{zheng2012ultralow}, S+A circuits from \cite{shiflett2020pixel}, 1-bit DAC from \cite{liu2019holylight}, ADC (8b, 1.2GSps) from \cite{adc}, Microcomb and optical spectral sharper from \cite{stern2018battery,xu202111}, EOM and On-chip SMF from \cite{xu202111}, MZIs from \cite{shen2017deep,gu2020towards}, and other periphery circuits from \cite{shafiee2016isaac}, respectively.

\begin{figure}[tb!]
\setlength{\abovecaptionskip}{0.05in}
\setlength{\belowcaptionskip}{-0.2in}
     \centering
     \includegraphics[width=0.43\textwidth]{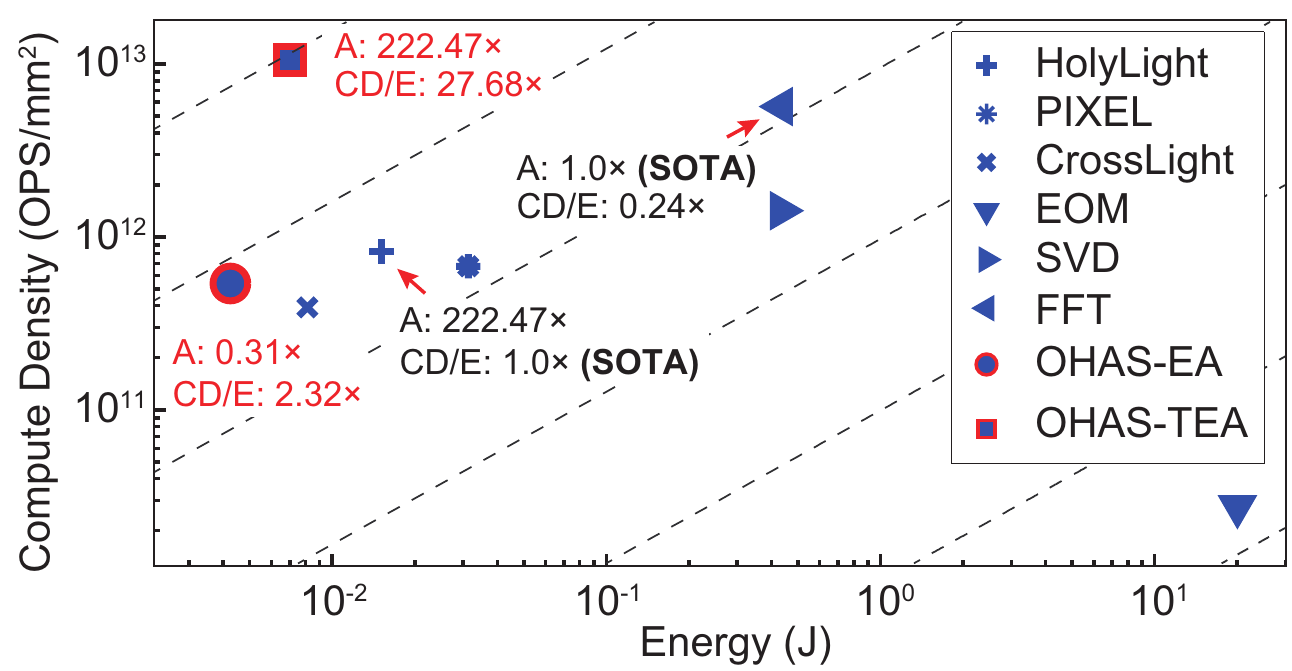}
     \caption{Average performance comparison of our O-HAS with the SOTA optical accelerators.}
     \label{f:OAs_avgPerf}
\end{figure}

\textbf{Accelerator performance.} Fig. \ref{f:OAs_avgPerf} illustrates the average performance of the generated OAs (OHAS-EA and OHAS-TEA) and the SOTA OAs. We adopt a figure-of-merit, Compute Density (CD), to describe OAs' throughput efficiency in OPS/$mm^2$, similar to \cite{nahmias2019photonic}. It shows that, among the six SOTA OAs, FFT achieves the highest compute density resulting from its excellent area efficiency, while HolyLight has the maximum throughput-per-energy. Compared with the SOTA OAs, OHAS-TEA improves the compute density by 1.86$\times$ and the throughput-per-energy by 27.68$\times$ with an area cost similar to HolyLight. Considering edge applications, which require OAs to have low energy and area costs), we also generate OHA-EA, which achieves minimized energy consumption and chip area cost with satisfying throughput: further improves the area efficiency by 69\% and achieves a 2.32$\times$ improvement in throughput-per-energy over SOTAs.

\begin{figure}[htbp]
\setlength{\abovecaptionskip}{0.05in}
\setlength{\belowcaptionskip}{-0.1in}
     \centering
     \includegraphics[width=0.5\textwidth]{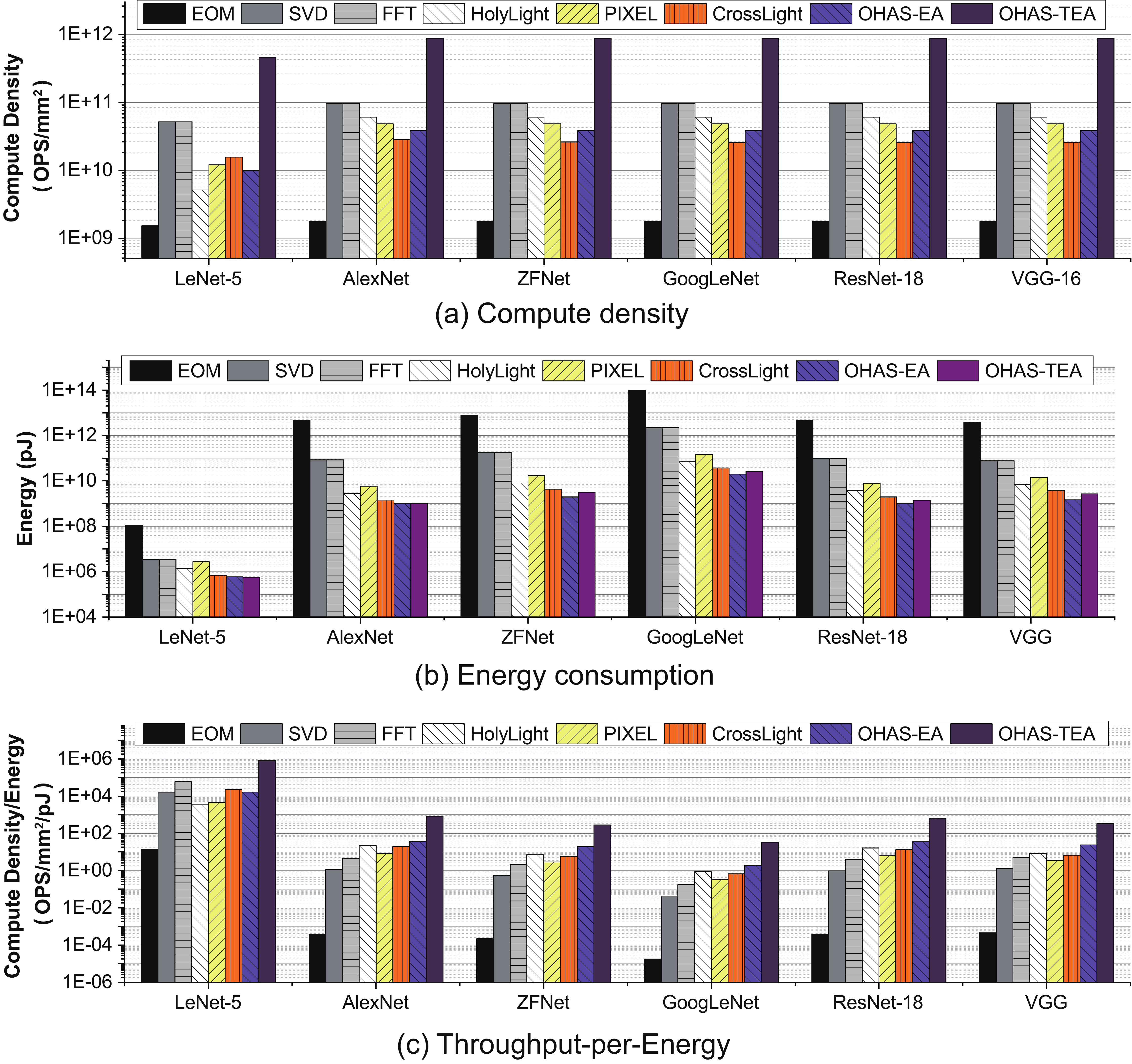}
     \caption{Performance comparison of our O-HAS with the SOTA optical accelerators on various DNN models.}
     \label{f:OHAS_perf}
\end{figure}

Fig. \ref{f:OHAS_perf} illustrates the performance of these hardware platforms on various DNN models, in terms of compute density, energy consumption, and throughput-per-energy. We can see that the performance of the OAs varies on different DNN modes and for the same DNN, different OA designs have distinct performance, from which we make three observations. First, our OHAS-TEA and OHAS-EA consistently outperform the SOTAs with greatly improved compute density and reduced energy cost.
Second, using analog computing, MZI-based OAs (i.e., SVD and FFT) in general achieve higher compute density. However, this advantage degrades on large DNNs due to their limited scalability. The computation reliability decreases as the size of MZI-based OAs increases, which will largely reduce the inference accuracy. Third, MR-based OAs excel in energy efficiency, e.g., HolyLight better favors large DNNs against PIXEL and CrossLight because it occupies the largest chip area (being composed of 28 tiles with 8 OCUs in each tile). Although EOM shows limited performance advantages over the others, it proposes a brand-new and promising wavelength-temporal-spatial interleaving methodology for OA design.
\begin{figure}[htbp]
\setlength{\abovecaptionskip}{0.05in}
\setlength{\belowcaptionskip}{-0.2in}
     \centering
     \includegraphics[width=0.5\textwidth]{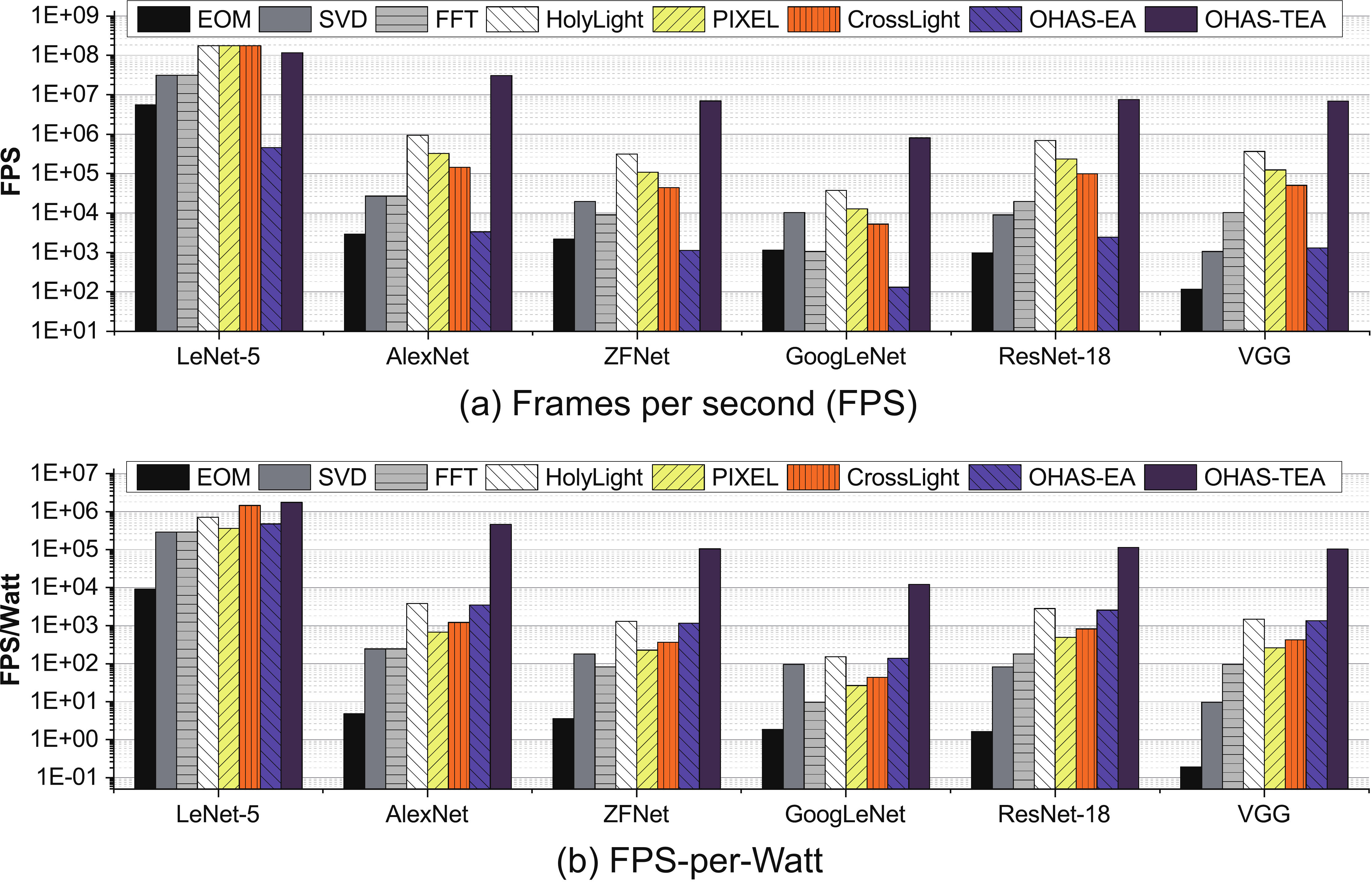}
     \caption{The inference performance of our O-HAS and the SOTA optical accelerators.}
     \label{f:OHAS_inf_perf}
\end{figure}

\textbf{Inference performance.} The DNN inference performance comparison of various accelerators is shown in Figure \ref{f:OHAS_inf_perf}. Our OHAS-TEA achieves both the highest frame-per-second (FPS) and performance-per-Watt for inference, with 0.67--32.48 $\times$ improvement in FPS and 1.20--119.70$\times$ improvement in FPS-per-Watt. Our OHAS-EA also achieves satisfactory performance comparable to the SOTA designs. For large DNNs, HolyLight achieves the best inference performance on large DNNs among all the SOTAs. However, on small DNNs like LeNet-5, CrossLight outperforms the others in FPS-per-Watt thanks to its energy benefits by exploiting efficient tuning techniques and an optimized MR design.
\begin{table}[b!]
\setlength{\abovecaptionskip}{0in}
\setlength{\belowcaptionskip}{-0.1in}
\centering
\footnotesize
\caption{Search efficiency of O-Search Engine.}
\setlength{\tabcolsep}{3pt}
\begin{tabular}{|c|c|c||c|c|c|}\hline
\textbf{DNN} &  \textbf{OA Space} & \textbf{\tabincell{c}{Search\\ Time (s)}} & \textbf{DNN} &  \textbf{OA Space} & \textbf{\tabincell{c}{Search\\ Time (s)}} \\\hline
LeNet-5  &  2.98E+05  & 9.76E+01 & ResNet-18 & 1.25E+06 & 5.12E+02 \\
AlexNet  &  4.76E+05 & 1.47E+02  & GoogLeNet & 3.45E+06 & 9E+03\\
ZFNet  &    4.17E+05 & 1.29E+02  & VGG-16 & 9.53E+05 & 3.29E+02\\\hline
\end{tabular}
\label{t:search_eff}
\end{table}

\textbf{Search efficiency.} We summarize the search space size and the search time for various DNNs in Tab. \ref{t:search_eff}, the latter is the time used by our O-Search Engine to generate OHAS-TEA for each DNN model. We can see that our O-Search Engine achieves a high search efficiency, and only require ten to thousand of seconds for various DNN models, while the performance of the generated OHAS-TEA is oustanding as reported above. There are two major contributors to the excellent search efficiency of O-HAS. First, our O-Cost Predictor possesses high prediction efficiency, as the analytical models proposed in Eq. (\ref{e:E_DNN})-(\ref{e:A_DNN}) all have a polynomial time complexity. Second, our O-Search Engine achieves a high and scalable search speed with an effective search strategy.
For example, the depths of GoogLeNet is about three times and even one order of magnitude larger than the other DNNs, for which our engine still achieved high search efficiency.

\begin{figure}[tb!]
\setlength{\abovecaptionskip}{0.1in}
\setlength{\belowcaptionskip}{-0.2in}
     \centering
     \includegraphics[width=0.46\textwidth]{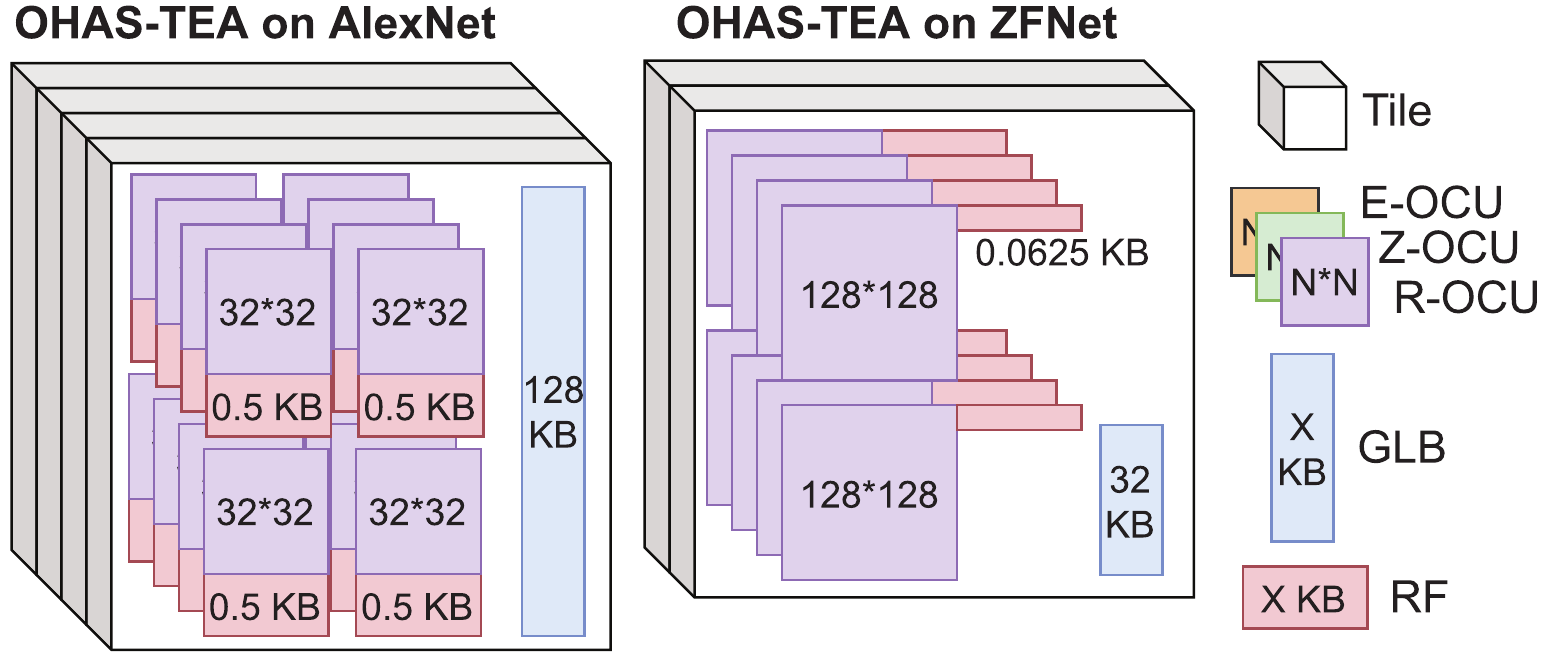}
     \caption{Visualization of the O-HAS generated accelerators on AlexNet and ZFNet.}
     \label{f:OHAS_vis}
\end{figure}

\textbf{Optical Accelerator Visualization.} To gain an intuitive understanding of the architecture of the O-HAS generated optical accelerators, Fig. \ref{f:OHAS_vis} visualizes the O-HAS generalized OAs (i.e., OHAS-TEA) on AlexNet and ZFNet, which achieve 119.7$\times$ and 81.82$\times$ improvements in FPS/Watt over the SOTA designs. AlexNet and ZFNet have a similar network structure but different network parameters. Accordingly, the O-HAS generated OAs are distinct.
\section{Conclusion}
In this paper, we have developed and evaluated O-HAS, the first framework for automated optical hardware accelerator search, aiming to boost both the acceleration performance and development speed of optical DNN accelerators. To realize fast and effective automated search, O-HAS consists of two integrated enablers. One is an accurate yet efficient O-Cost Predictor, which is able to predict an optical accelerator's energy, latency, and area costs based on the given DNN model and optical accelerator design. The other is an O-Search Engine, which automatically explores the huge design space of optical DNN accelerators and generates the optimal accelerators for the target acceleration performance and efficiency. Extensive experiments and ablation studies consistently validate the effectiveness of our O-HAS framework, with an improvement of up to 119.7$\times$ over the state-of-the-art designs.

\section*{Acknowledgment}
This work is partially supported by the National Science Foundation CAREER program (Award number: 2048183) and CCRI program (Award number: 2016727).

\bibliographystyle{IEEEtran}
\bibliography{Main}

\begin{thebibliography}{10}
\providecommand{\url}[1]{#1}
\csname url@samestyle\endcsname
\providecommand{\newblock}{\relax}
\providecommand{\bibinfo}[2]{#2}
\providecommand{\BIBentrySTDinterwordspacing}{\spaceskip=0pt\relax}
\providecommand{\BIBentryALTinterwordstretchfactor}{4}
\providecommand{\BIBentryALTinterwordspacing}{\spaceskip=\fontdimen2\font plus
\BIBentryALTinterwordstretchfactor\fontdimen3\font minus
  \fontdimen4\font\relax}
\providecommand{\BIBforeignlanguage}[2]{{%
\expandafter\ifx\csname l@#1\endcsname\relax
\typeout{** WARNING: IEEEtran.bst: No hyphenation pattern has been}%
\typeout{** loaded for the language `#1'. Using the pattern for}%
\typeout{** the default language instead.}%
\else
\language=\csname l@#1\endcsname
\fi
#2}}
\providecommand{\BIBdecl}{\relax}
\BIBdecl

\bibitem{chen2016eyeriss}
Y.-H. Chen, T.~Krishna, J.~S. Emer, and V.~Sze, ``Eyeriss: An energy-efficient
  reconfigurable accelerator for deep convolutional neural networks,''
  \emph{IEEE journal of solid-state circuits}, vol.~52, no.~1, pp. 127--138,
  2016.

\bibitem{zhao2020smartexchange}
Y.~Zhao, X.~Chen, Y.~Wang, C.~Li, H.~You, Y.~Fu, Y.~Xie, Z.~Wang, and Y.~Lin,
  ``Smartexchange: Trading higher-cost memory storage/access for lower-cost
  computation,'' 2020.

\bibitem{9138916}
W.~Li, P.~Xu, Y.~Zhao, H.~Li, Y.~Xie, and Y.~Lin, ``Timely: Pushing data
  movements and interfaces in pim accelerators towards local and in time
  domain,'' in \emph{2020 ACM/IEEE 47th Annual International Symposium on
  Computer Architecture (ISCA)}, 2020, pp. 832--845.

\bibitem{nahmias2019photonic}
M.~A. Nahmias, T.~F. De~Lima, A.~N. Tait, H.-T. Peng, B.~J. Shastri, and P.~R.
  Prucnal, ``Photonic multiply-accumulate operations for neural networks,''
  \emph{IEEE Journal of Selected Topics in Quantum Electronics}, vol.~26,
  no.~1, pp. 1--18, 2019.

\bibitem{shastri2021photonics}
B.~J. Shastri, A.~N. Tait, T.~F. de~Lima, W.~H. Pernice, H.~Bhaskaran, C.~D.
  Wright, and P.~R. Prucnal, ``Photonics for artificial intelligence and
  neuromorphic computing,'' \emph{Nature Photonics}, vol.~15, no.~2, pp.
  102--114, 2021.

\bibitem{shen2017deep}
Y.~Shen, N.~C. Harris, S.~Skirlo, M.~Prabhu, T.~Baehr-Jones, M.~Hochberg,
  X.~Sun, S.~Zhao, H.~Larochelle, D.~Englund \emph{et~al.}, ``Deep learning
  with coherent nanophotonic circuits,'' \emph{Nature Photonics}, vol.~11,
  no.~7, p. 441, 2017.

\bibitem{liu2019holylight}
W.~Liu, W.~Liu, Y.~Ye, Q.~Lou, Y.~Xie, and L.~Jiang, ``Holylight: A
  nanophotonic accelerator for deep learning in data centers,'' in \emph{2019
  Design, Automation \& Test in Europe Conference \& Exhibition (DATE)}.\hskip
  1em plus 0.5em minus 0.4em\relax IEEE, 2019, pp. 1483--1488.

\bibitem{xu202111}
X.~Xu, M.~Tan, B.~Corcoran, J.~Wu, A.~Boes, T.~G. Nguyen, S.~T. Chu, B.~E.
  Little, D.~G. Hicks, R.~Morandotti \emph{et~al.}, ``11 tops photonic
  convolutional accelerator for optical neural networks,'' \emph{Nature}, vol.
  589, no. 7840, pp. 44--51, 2021.

\bibitem{shiflett2020pixel}
K.~Shiflett, D.~Wright, A.~Karanth, and A.~Louri, ``Pixel: Photonic neural
  network accelerator,'' in \emph{2020 IEEE International Symposium on High
  Performance Computer Architecture (HPCA)}.\hskip 1em plus 0.5em minus
  0.4em\relax IEEE, 2020, pp. 474--487.

\bibitem{sunny2021crosslight}
F.~Sunny, A.~Mirza, M.~Nikdast, and S.~Pasricha, ``Crosslight: A cross-layer
  optimized silicon photonic neural network accelerator,'' 2021.

\bibitem{shafiee2016isaac}
A.~Shafiee, A.~Nag, N.~Muralimanohar, R.~Balasubramonian, J.~P. Strachan,
  M.~Hu, R.~S. Williams, and V.~Srikumar, ``Isaac: A convolutional neural
  network accelerator with in-situ analog arithmetic in crossbars,'' \emph{ACM
  SIGARCH Computer Architecture News}, vol.~44, no.~3, pp. 14--26, 2016.

\bibitem{gu2020towards}
J.~Gu, Z.~Zhao, C.~Feng, Z.~Ying, M.~Liu, R.~T. Chen, and D.~Z. Pan, ``Towards
  hardware-efficient optical neural networks: Beyond fft architecture via joint
  learnability,'' \emph{IEEE Transactions on Computer-Aided Design of
  Integrated Circuits and Systems}, 2020.

\bibitem{gusqueezelight}
J.~Gu, C.~Feng, Z.~Zhao, Z.~Ying, M.~Liu, R.~T. Chen, and D.~Z. Pan,
  ``Squeezelight: Towards scalable optical neural networks with multi-operand
  ring resonators.''

\bibitem{wetzstein2020inference}
G.~Wetzstein, A.~Ozcan, S.~Gigan, S.~Fan, D.~Englund, M.~Solja{\v{c}}i{\'c},
  C.~Denz, D.~A. Miller, and D.~Psaltis, ``Inference in artificial intelligence
  with deep optics and photonics,'' \emph{Nature}, vol. 588, no. 7836, pp.
  39--47, 2020.

\bibitem{parashar2019timeloop}
A.~Parashar, P.~Raina, Y.~S. Shao, Y.-H. Chen, V.~A. Ying, A.~Mukkara,
  R.~Venkatesan, B.~Khailany, S.~W. Keckler, and J.~Emer, ``Timeloop: A
  systematic approach to dnn accelerator evaluation,'' in \emph{2019 IEEE
  international symposium on performance analysis of systems and software
  (ISPASS)}.\hskip 1em plus 0.5em minus 0.4em\relax IEEE, 2019, pp. 304--315.

\bibitem{fodlam}
A.~Sampson and M.~Buckler, ``{FODLAM}, a first-order deep learning accelerator
  model,'' \url{https://github.com/cucapra/fodlam}.

\bibitem{gumbel1954statistical}
E.~J. Gumbel, \emph{Statistical theory of extreme values and some practical
  applications: a series of lectures}.\hskip 1em plus 0.5em minus 0.4em\relax
  US Government Printing Office, 1954, vol.~33.

\bibitem{maddison2014sampling}
C.~J. Maddison, D.~Tarlow, and T.~Minka, ``A* sampling,'' \emph{arXiv preprint
  arXiv:1411.0030}, 2014.

\bibitem{thop}
``{THOP}: Pytorch-opcounter,''
  \\\url{https://github.com/Lyken17/pytorch-OpCounter}.

\bibitem{kingma2014adam}
D.~P. Kingma and J.~Ba, ``Adam: A method for stochastic optimization,''
  \emph{arXiv preprint arXiv:1412.6980}, 2014.

\bibitem{stern2018battery}
B.~Stern, X.~Ji, Y.~Okawachi, A.~L. Gaeta, and M.~Lipson, ``Battery-operated
  integrated frequency comb generator,'' \emph{Nature}, vol. 562, no. 7727, pp.
  401--405, 2018.

\bibitem{zheng2012ultralow}
X.~Zheng, F.~Y. Liu, J.~Lexau, D.~Patil, G.~Li, Y.~Luo, H.~D. Thacker,
  I.~Shubin, J.~Yao, K.~Raj \emph{et~al.}, ``Ultralow power 80 gb/s arrayed
  cmos silicon photonic transceivers for wdm optical links,'' \emph{Journal of
  Lightwave Technology}, vol.~30, no.~4, pp. 641--650, 2012.

\bibitem{subbaraman2015recent}
H.~Subbaraman, X.~Xu, A.~Hosseini, X.~Zhang, Y.~Zhang, D.~Kwong, and R.~T.
  Chen, ``Recent advances in silicon-based passive and active optical
  interconnects,'' \emph{Optics express}, vol.~23, no.~3, pp. 2487--2511, 2015.

\bibitem{hosseini20111}
A.~Hosseini, D.~N. Kwong, Y.~Zhang, H.~Subbaraman, X.~Xu, and R.~T. Chen,
  ``1$\times$ n multimode interference beam splitter design techniques for
  on-chip optical interconnections,'' \emph{IEEE Journal of Selected Topics in
  Quantum Electronics}, vol.~17, no.~3, pp. 510--515, 2011.

\bibitem{adc}
B.~Murmann, ``{ADC} performance survey 1997-2021,''
  \url{http://web.stanford.edu/~murmann/adcsurvey.html}.

\end{thebibliography}

\end{document}